\documentclass[10pt,twocolumn,letterpaper]{article}

\usepackage[pagenumbers]{cvpr} 
\definecolor{cvprblue}{rgb}{0.21,0.49,0.74}
\usepackage[pagebackref,breaklinks,colorlinks,allcolors=cvprblue]{hyperref}
\usepackage{stfloats}

\def\paperID{10586} 
\def\confName{CVPR}
\def\confYear{2025}

\title{Enhancing SAM with Efficient Prompting and Preference Optimization for Semi-supervised Medical Image Segmentation}

\newcommand*\samethanks[1][\value{footnote}]{\footnotemark[#1]}
\author{
    Aishik Konwer\textsuperscript{1}\thanks{The work was done during an internship at GE Healthcare.}, 
    Zhijian Yang\textsuperscript{2}\thanks{Corresponding authors: \{erhan.bas, zhijian.yang\}@gehealthcare.com}, 
    Erhan Bas\textsuperscript{2}\samethanks, 
    Cao Xiao\textsuperscript{2}, 
    Prateek Prasanna\textsuperscript{1},\\
    Parminder Bhatia\textsuperscript{2},
    Taha Kass-Hout\textsuperscript{2}\\
    \textsuperscript{1}Stony Brook University \\
    \textsuperscript{2}GE Healthcare
}

\begin{document}
\maketitle
\begin{abstract}
Foundational models such as the Segment Anything Model~(SAM) are gaining traction in medical imaging segmentation, supporting multiple downstream tasks. However, such models are supervised in nature, still relying on large annotated datasets or prompts supplied by experts. Conventional techniques such as active learning to alleviate such limitations are limited in scope and still necessitate continuous human involvement and complex domain knowledge for label refinement or establishing reward ground truth. To address these challenges, we propose an enhanced Segment Anything Model (SAM) framework that utilizes annotation-efficient prompts generated in a fully unsupervised fashion, while still capturing essential semantic, location, and shape information through contrastive language-image pretraining and visual question answering. We adopt the direct preference optimization technique to design an optimal policy that enables the model to generate high-fidelity segmentations with simple ratings or rankings provided by a virtual annotator simulating the human annotation process. State-of-the-art performance of our framework in tasks such as lung segmentation, breast tumor segmentation, and organ segmentation across various modalities, including X-ray, ultrasound, and abdominal CT, justifies its effectiveness in low-annotation data scenarios.
\end{abstract}    
\section{Introduction}
\label{sec:intro}

With advancements in medical image analysis, there is an increasing need for sophisticated methods~\cite{survey} to leverage the vast availability of radiology datasets (such as X-ray, CT, and MRI) for accurate organ and tumor segmentation, as well as disease classification. The results of these tasks are crucial for physicians in designing effective treatment plans and surgical procedures. Several state-of-the-art deep learning-based foundational models, such as Vision-Language Models (VLMs), are now available for these purposes, relying on custom prompting to generate relevant predictions. However, they face two significant challenges: (1) despite being trained with only sparse prompts such as points or bounding boxes, these models still require human supervision for the prompt generation, leading to inefficiencies; and (2) many datasets lack comprehensive annotations, resulting in under-utilization during training of complex, data-hungry foundational architectures. Additionally, the high cost of human annotation efforts to create ground truth data can significantly escalate model development expenses.

\begin{figure*}[t]
  \centering
  \includegraphics[width=0.8\linewidth]{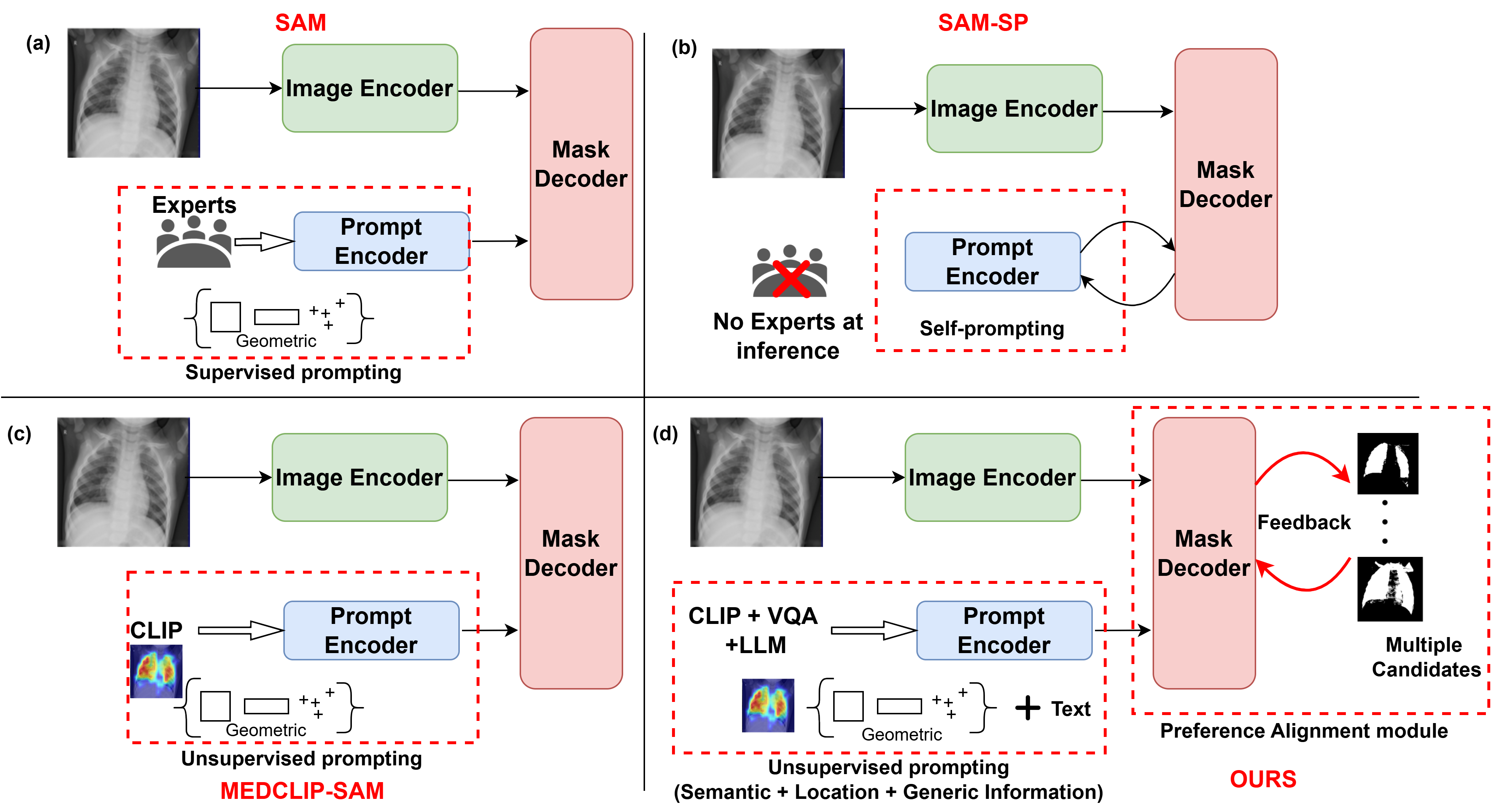}
    \caption{\textbf{Overview of our model:}
\textbf{(a)} SAM and SAM-based approaches rely on expert prompts during both training and inference. \textbf{(b)} SAM-SP~\cite{samsp} introduces a self-promoting module, eliminating the need for expert prompts during inference.
\textbf{(c)} MedCLIP-SAM~\cite{medclipsam} uses unsupervised semantic prompts to generate pseudo-labels through SAM.
\textbf{(d)} Our approach not only combines semantic, location, and generic information via unsupervised prompts but also introduces a preference-based alignment module to reward or penalize the model. 
}
    \label{teaser}
\end{figure*}

Popular architectures such as SAM and CLIP, have been extended to medical data, and led to innovations such as BiomedCLIP, Merlin, SAM-Med2D, and SAM-Med3D~\cite{biomedclip, merlin,sammed2d,sammed3d}. Recently, numerous studies have focused on enhancing SAM by replacing geometric prompts and integrating semantic and spatial knowledge in an unsupervised manner through techniques such as (1) self-prompting, (2) class activation maps from CLIP, and (3) object localization models~\cite{samsp, medclipsam, medlsam}. As shown in Fig.~\ref{teaser}, self-promoting (b) does not require expert-provided prompts during inference. On the other hand, unsupervised prompts (c) in~\cite{medclipsam} are only used for training SAM to generate pseudo-labels for weakly supervised downstream tasks. They lack sufficient location and shape information, which could be provided by text-based prompts. Hence, the question arises: \textit{Can we come up with more refined and efficient prompts that can deliver stronger signals to foundational models without requiring intervention from annotators in both training and testing stages?} State-of-the-art models also lack comprehensive integration of semantic, locational, and generic class information in their prompting strategies. In our approach, we leverage CLIP, VQA, and LLM models~\cite{biomedclip,pmcvqa,gpt4} to extract this discriminative information, improving segmentation performance in unsupervised settings.

To tackle the challenges posed by limited annotated datasets, several weakly supervised semantic segmentation algorithms~\cite{singlestage,medclipsam} have been proposed. Single-stage methods~\cite{singlestage} leverage coarse image labels to perform segmentation in an end-to-end fashion, while many two-stage approaches~\cite{medclipsam} utilize foundational models to generate pseudo labels that can be used to train downstream segmentation architectures. Additionally, human-in-the-loop frameworks~\cite{alignnatural} have proven beneficial for tasks such as image generation, segmentation, and prognosis, involving annotators to refine segmentation labels or assess the plausibility of synthetic images. This feedback can either be incorporated back into the model or used as ground truth to train a reward function via reinforcement learning (RL), significantly reducing the need for annotated training data. However, these methods still rely on explicit reward modeling, which prevents end-to-end training. In some cases, annotators must provide complex ratings or refinements, making it difficult to train a reward function efficiently. 
We ask another question: \textit{Can we skip training a reward function and develop a straightforward, end-to-end pipeline that relies on simple annotator preferences?}
To address this, we draw inspiration from direct preference optimization techniques in the LLM preference alignment literature. We propose a preference-based reward model within our framework, using a novel loss function for end-to-end training. Our approach thus aligns with human preferences without the need for explicit reward modeling and is both easy to implement and train.

Our first goal is to develop refined and efficient prompts to fine-tune the SAM-Med2D encoder on diverse datasets, including 2D chest X-rays, breast ultrasound, and 3D abdominal CT scans. Initially, we input an image into the encoder alongside the BiomedCLIP~\cite{biomedclip} and MedVInT (VQA)~\cite{pmcvqa} models. Corresponding texts, examples such as ``Chest X-ray" and ``Describe the condition of the lungs and location of pathologies," are fed into BiomedCLIP and MedVInT, respectively. We gather generic information for the disease class from GPT-4, which is combined with the answers from the VQA model. Saliency maps generated by the CLIP model undergo dense CRF postprocessing to obtain bounding boxes. These bounding boxes and textual prompts are then inputted into the prompt encoder. Subsequently, the mask decoder receives both the image encoder embeddings and prompt embeddings to produce the segmentation maps.

After fine-tuning the prompting module on a small proportion of annotated data, we introduce our second major idea: simulating human feedback through a AI preference alignment module. We generate multiple segmentation candidates for a given image by thresholding the SAM output probabilities at various levels. These candidates are rated on a scale of 1 to 4 based on the overlap between the candidates and ground truth, mimicking the evaluation process of a human annotator. This approach does not require explicit access to ground truth data for training, thus making our task a form of semi-supervised segmentation. We propose a DPO-inspired~\cite{dpo} loss function that encourages the model to prioritize desirable segmentation outputs by rewarding higher-rated candidates and penalizing lower-rated ones. The model is thus trained on the remaining portion of the dataset, without annotations, to perform relevant medical image segmentation tasks. 

Overall our contributions can be summarized as follows:
\begin{itemize}
    \item We propose refined and efficient unsupervised prompting strategies that deliver comprehensive information about location, semantics, and general disease/organ characteristics to our SAM-Med2D-based framework. Such an approach enhances segmentation performance while reducing reliance on human input for geometric prompts.

    \item We introduce a novel DPO-inspired loss function that facilitates semi-supervised model improvement using simulated human-in-the-loop feedback, eliminating the need for an explicit reward function. The framework generates multiple segmentation maps and rates them based on segmentation overlap, mimicking the evaluation of a human annotator. The model learns to distinguish between favorable and unfavorable candidates effectively.
\end{itemize}

\section{Related Work}
\subsection{Vision-language models for medical domain}
CLIP~\cite{clip} has gained much popularity in medical image analysis.~\cite{pubmedclip} fine-tuned CLIP on various PubMed articles to create PubMedCLIP. MedCLIP~\cite{medclip} leverages unpaired image and text datasets along with a semantic-matching loss to align visual and textual information. Windsor et al.~\cite{cliplowdata} employ unimodal self-supervision, local-global contrastive losses, and data augmentation to enhance zero-shot performance and retrieval task efficiency in low-resource data settings. Some modality-specific CLIP variants~\cite{cxrclip,ghosh2024mammo} have been developed for Chest X-ray and Mammograms due to the readily available image-text data in these areas. However, BiomedCLIP~\cite{biomedclip} stands out as the most recent model, excelling in scalability and performance across diverse multi-organ cross-modal retrieval tasks. Therefore, we have integrated BiomedCLIP into the CLIP-driven bounding box generation module of our framework.

Building on the success of large language models (LLMs) such as LLaMa~\cite{llama} and GPT~\cite{gpt}, researchers have explored merging visual features with textual representations using techniques such as cross-attention, Q-former, instruction tuning, and projection layers. This effort has resulted in vision-language models (VLMs)~\cite{flamingo,blip2,llava,minigpt4}, including Flamingo, BLIP-2, LLaVA, and MiniGPT, which were further adapted for the medical domain through pretraining on multimodal medical datasets. Med-Flamingo~\cite{medflamingo} is the first medical visual question-answering (VQA) model with few-shot generation capabilities. RadFM~\cite{radfm} serves as a foundational model that also incorporates 3D volume data. Pretraining on extensive datasets, PMC-15M and PMC-VQA~\cite{biomedclip,pmcvqa}, has facilitated the development of LLaVA-Med~\cite{llavamed} and MedVInT~\cite{pmcvqa}, respectively. Our segmentation pipeline leverages MedVInT’s capabilities for enhanced localization and shape-based answer generation related to tumors, organs, and disease manifestations.

\subsection{SAM for multimodal biomedical data}
MedSAM and SAM-Med2D~\cite{medsam,sammed2d} focused on adapting SAM~\cite{sam} for medical applications by fine-tuning it on 2D medical datasets, while SAM-Med3D~\cite{sammed3d} introduced alternative modules to accommodate 3D volumes. Efficient approaches, for example, AutoSAM~\cite{autosam} utilize trainable prompt encoders, whereas FastSAM3D~\cite{fastsam3d} employs flash attention to accelerate inference. MedLSAM~\cite{medlsam} proposed a localization framework to generate 3D bounding boxes as prompt input. However, most of these methods require ground truth data (bounding boxes or points) for training the prompt encoder, whereas we propose an unsupervised route for the same. We leverage ad-hoc VLM models such as CLIP, VQA, and GPT-4 together to propagate comprehensive information—encompassing semantics, location, and generic disease/organ information—that significantly enhances segmentation performance.

\subsection{Human-in-the-loop feedback}
Human-in-the-loop training paradigm in medical imaging has primarily focused on two areas: active learning~\cite{active}, which identifies optimal data points for labeling to maximize performance, and interactive feedback~\cite{interactive} on model predictions to calibrate parameters. Examples include UI-Net, DeepIGeoS, and BIFseg~\cite{uinet,deepigeos,bifseg}, which utilize expert-provided scribbles or bounding boxes alongside geodesic transforms and graph-cut techniques to refine segmentation labels. Rao et al.~\cite{imil} introduced IMIL, the first framework that assigns clinicians to guide data augmentations on mispredictions, emphasizing disease-relevant regions while eliminating irrelevant ones. Recently, the success of human feedback in instruction-tuning, and aligning large language model outputs through RL objective ~\cite{lang1,lang2}, has led researchers to utilize human preferences to evaluate synthetic natural and histopathology images, thereby improving image-to-text models~\cite{alignnatural,alignpath} by training a reward function. Training a reward function requires the curation of dedicated human preference datasets and also prohibits the framework from operating in an end-to-end manner. Additionally, it often demands advanced domain knowledge from annotators, which can be costly. Our approach addresses this challenge by using direct preference optimization~\cite{dpo} to fulfill the RL objective, enabling the framework to serve as its own reward model. The model aligns generated proposals with appropriate preferences or ratings, leading to performance improvements, even with limited annotated data. These preferences are generated through a simulation mimicking human-in-the-loop feedback. Crucially, our method is easier to train and implement than traditional reward function-based RL pipelines.

\begin{figure*}[hbtp]
  \centering
  \includegraphics[width=0.8\linewidth]{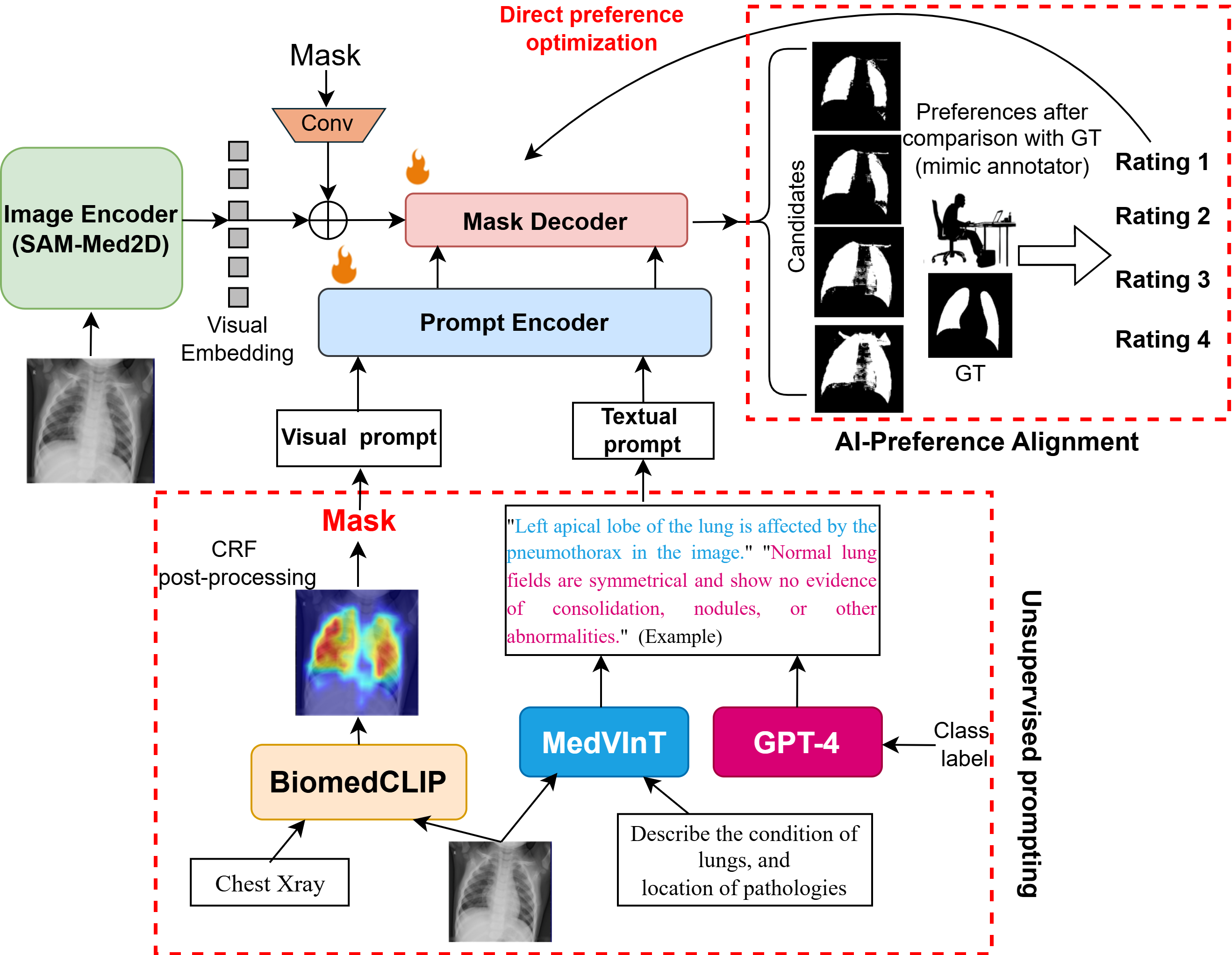}
    \caption{\textbf{Illustration of the proposed framework for semi-supervised segmentation:} Unsupervised geometric and text prompts, obtained from pretrained BiomedCLIP, MedVInT, and GPT-4 models, are fed into the prompt encoder for finetuning the framework on a small fraction of annotated data. In the next stage, we simulate a virtual annotation process that assigns ratings to the generated segmentation candidates, which are used to fine-tune the decoder. This stage handles unannotated data, as the model does not rely on ground truth for direct supervision but only for rating while simulating a human annotator's feedback.}
    \label{framework}
\end{figure*}


\section{Methodology}
\textbf{Overview.} Given an input image, our aim is to generate prompts in the form of bounding boxes and textual queries, which will be used to produce segmentation masks. 
The image is processed by three pretrained components: the SAM-Med2D encoder, BiomedCLIP, and MedVInT. Concurrently, GPT-4 curates generic information about the concerned disease or organ. 
This information, combined with the BiomedCLIP-generated box prompt and the MedVInT-provided textual answer prompt, serves as input for the prompt encoder. The prompt generation process is detailed in Sec.~\ref{section:prompt}. Next, both the encoded image and prompt embeddings are fed into the decoder to create segmentation maps. Following this initial fine-tuning on annotated data, we introduce a novel loss to mimic human-in-the-loop feedback. By thresholding the output probabilities, our model produces four different segmentation candidates, and we incorporate insights about the quality of these candidates into the framework through the proposed loss function. This part, trained with unannotated data, is elaborated in Sec.~\ref{section:feedback}.

\subsection{Visual and textual prompt generation}\label{section:prompt}
\textbf{Visual prompt.} To generate visual prompts, we first input the image along with corresponding text prompts (e.g. ``chest x-ray", ``benign breast tumor", ``left kidney", ``liver", etc.) into BiomedCLIP, a foundational model pretrained on millions of medical image-caption pairs. Next, we leverage gScoreCAM~\cite{gscore} to create a saliency map highlighting targeted regions (organs or tumors) in the image corresponding to the supplied text. These saliency maps are then post-processed with a conditional random field filter~\cite{crf} to produce coarse segmentation masks. We apply an area constraint, retaining the largest connected component or up to two, depending on the dataset (the lung dataset may yield two components). Finally, we identify the bounding box coordinates within these closed components for the box prompt, while randomly sampling several points from the designated area for the point prompt. 

\textbf{Textual prompt.} We extract visual embeddings from a given image by processing it through a vision encoder derived from the PMC-CLIP architecture~\cite{pmcclip}. This encoder features a pretrained ResNet50~\cite{resnet} and a trainable projection layer constructed with stacks of transformer decoder blocks. Next, we create a prompt template that incorporates the question for the image, formatted as ``Question: \{\}, Answer is:". This prompt is then passed through a tokenization embedding layer initialized with the weights of PMC-LLaMA~\cite{pmcllama} to generate the text embedding. Finally, we concatenate the visual and text embeddings to form the input space for a pre-initialized multimodal transformer decoder. The answer generated from this VQA setup delivers essential information regarding the shape and location of anatomical structures and pathologies. Sample questions included in the prompt are: ``What is the shape of the liver and where is it located?", and ``What is the shape of breast tumor and where is it located?" Additionally, we prompt GPT-4 with the relevant organ or disease label to obtain a generic description of its characteristics. Finally, both types of textual prompts are concatenated and provided as input to the prompt encoder, described in the following paragraph.

\textbf{Prompt encoder and mask decoder.}
The prompt encoder, same as the one in SAM, supports three types of prompts: point, box, and text. Point and box prompts are represented by their positional encodings—specific coordinates for points, and the top-left and bottom-right corners for boxes—along with learnable feature representations. Text prompts are processed through a pretrained BiomedCLIP encoder to generate corresponding text embeddings. All prompt embeddings are then projected into 256-dimensional vectors. The feature map from the first iteration of the model, is downsampled through multiple convolutional layers followed by GELU activation to match the 256-channel dimension. Finally, these downsampled masks are combined element-wise with visual encoder emeddings. The mask decoder takes both the prompt embeddings and the visual embeddings to produce a segmentation map. The architecture is illustrated in Fig.~\ref{framework}.

\subsection{Segmentation proposal generation and Preference Alignment}\label{section:feedback}
\textbf{Proposal generation.} After fine-tuning our framework,comprising the prompt encoder and mask decoder, using ground truth for a fraction of the dataset, we propose integrating AI-based preferences for the next training episode. First, we generate multiple segmentation proposals by sampling different thresholds from the pixel probability scores output by the model. We then simulate an expert annotator to evaluate the quality of these proposals, assigning ratings based on the overlap between each generated proposal and the corresponding ground truth. Although ground truth is not explicitly used in this training phase, it plays a passive role by enabling the simulation of preferences in the absence of an actual human annotator in our experiments. We also experiment with two alternative rating mechanisms: one inspired by SAM, which backpropagates loss solely for the best candidate, and another that ranks all proposals rather than merely rating them. After obtaining the ratings or rankings, we fine-tune our decoder using the direct preference optimization technique to better align the segmentation outputs with the preferences of the virtual annotator.

\textbf{RLHF Preliminaries.} 
Language models typically utilize a reward function to align their generated responses with user preferences. This preference distribution is often modeled using the Bradley-Terry~\cite{bradley} model, which operates on pairwise comparisons. For a given prompt $X$, when the language model produces two responses, one more favorable, 
$Y_m$, and the other less favorable, $Y_l$, the distribution can be expressed as:
\begin{equation}
    P(Y_m>Y_l|X)=\frac{e^{{r^*}(X,Y_m)}}{e^{{r^*}(X,Y_m)} + e^{{r^*}(X,Y_l)}},
\end{equation}

Parameters of the reward model can be estimated via Maximum Likelihood Estimation (MLE) and the goal is to minimize the below loss function:
\begin{equation}\label{eq:reward_model}
    \mathcal{L_R} = -\mathbb{E}_{(X, Y_m, Y_l)\sim \mathcal{D}}\bigl[\log \sigma(r_{\phi}(X, Y_m)- r_{\phi}(X, Y_l))\bigr],
\end{equation}
where $\sigma$ is the logistic function and $\mathcal{D}$ denotes the dataset of preferences.

Reinforcement Learning from Human Feedback (RLHF) is a widely used method that involves training a reward model based on user ratings. The primary goal is to discover an optimal policy with parameter $\pi_{\theta}$ that maximizes this reward function while incorporating a KL divergence term to ensure the model's outputs do not deviate significantly from the original policy $\pi_{ref}$. The optimization problem can be formulated as follows:

\begin{equation}\label{eq:RL}
\begin{gathered}
    \max_{\pi_{\theta}}  \mathbb{E}_{X\sim \mathcal{D}, Y\sim \pi_{\theta}(Y \mid X)}\bigl[r_{\phi}(X, Y)\bigr] \\
- \beta\mathbb{D}_{\textrm{KL}}\bigl[\pi_{\theta}(Y\mid X)\mid \mid \pi_{ref}(Y\mid X)\bigr],
\end{gathered}
\end{equation}

\textbf {Direct Preference optimization.} Due to the discrete nature of language generation, this objective is not differentiable, which usually necessitates optimization through reinforcement learning. The primary language model then leverages this framework to align its outputs with user ratings, ultimately generating high-scoring responses. In contrast, Direct Preference Optimization (DPO) is a more recent and streamlined approach that eliminates the need for a separate reward model. Instead, it fine-tunes the main language model directly using user preferences. This is achievable because DPO focuses on optimizing the policy itself rather than the reward function.
The maximum likelihood objective for a parameterized policy $\pi_{\theta}$ in such a case can be formulated as:

\begin{equation}\label{eq:optimum_model}
\scriptsize
\begin{gathered}
    \mathcal{L}_\text{DPO}(\pi_{\theta}; \pi_{ref}) = -\mathbb{E}_{(X, Y_m, Y_l)\sim \mathcal{D}}\bigg[\log \sigma \bigg(\beta \log \frac{\pi_{\theta}(Y_m\mid X)}{\pi_{ref}(Y_m\mid X)} \\
    - \beta \log \frac{\pi_{\theta}(Y_l\mid X)}{\pi_{ref}(Y_l\mid X)}\bigg)\bigg],
\end{gathered}
\end{equation}

where $\beta$ is the weight applied to reward or penalize the responses.
In DPO, models are fine-tuned using pairs of outputs, explicitly comparing a preferred response \( Y_m \) with less preferred ones \( Y_l \). We adapt this concept for our application in the imaging domain, where multiple segmentation candidates \( Y_1, ..., Y_4 \) are generated for the same \{image, prompt\} pair \( I \). With ratings available for each candidate, we assign weights $\beta_1, \beta_2$ to these candidates (from best to worst) to reward the higher-quality segmentations \( Y_1, Y_2 \) and penalize the less desirable ones \( Y_3, Y_4 \). Consequently, Equation \ref{eq:optimum_model} is modified to create Equation \ref{eq:final_model}. This approach allows us to incorporate real-world annotator preferences into our method through a straightforward yet effective loss function. The parameters of the initially fine-tuned model from Sec. \ref{section:prompt} are denoted as \( \pi_{\text{fine}} \). Our objective is to find the optimal parameters \( \pi_{\psi} \) for the final architecture without deviating much from \( \pi_{\text{fine}} \).

\vspace{-0.1in}
\begin{equation}\label{eq:final_model}
\scriptsize
\begin{gathered}
    \mathcal{L}_\text{DPO}(\pi_{\psi}; \pi_{fine}) = -\mathbb{E}_{(I, Y_1, Y_2, Y_3, Y_4)\sim \mathcal{D}}\bigg[\log \sigma \big(\beta_1 \log \frac{\pi_{\psi}(Y_1\mid I)}{\pi_{fine}(Y_2\mid I)} \\ 
    + \beta_2 \log \frac{\pi_{\psi}(Y_2\mid I)}{\pi_{fine}(Y_2\mid I)} 
    - \beta_2 \log \frac{\pi_{\psi}(Y_3\mid I)}{\pi_{fine}(Y_3\mid I)} 
    - \beta_1 \log \frac{\pi_{\psi}(Y_4\mid I)}{\pi_{fine}(Y_4\mid I)}\big)\bigg]
\end{gathered}
\end{equation}

\section{Experimental Design and Results}

\textbf{Datasets.} We evaluated our framework for semi-supervised segmentation using three public datasets, covering lung, breast tumor, and abdominal organ segmentation tasks across multiple radiology modalities, including X-ray, ultrasound, and CT, as detailed below:

\textbf{Ultrasound Breast Tumor segmentation:} The dataset consists of 810 images, combining cases from the Breast Ultrasound Images (BUSI)~\cite{busi} dataset (437 benign, 210 malignant) and the UDIAT dataset~\cite{udiat} (109 benign, 54 malignant). Of these, 600 images were used for training and 210 for testing.

\textbf{Chest X-ray Lung Segmentation:}
We used 27,132 chest X-ray images from the COVID-19 Radiography Database (COVID-QU-Ex)~\cite{covid}, including images labeled normal, lung opacity, viral pneumonia, and covid for training the model. A separate set of 6,788 images was used for testing.

\textbf{Abdominal CT Organ Segmentation:}
For segmentation of 15 different abdominal organs, we utilized all 200 annotated CT scans from the training set of the AMOS-CT dataset~\cite{amos}. Model evaluations were conducted on the 100 CT scans from the validation set.

\textbf{Implementation Details.}
Our method is implemented in PyTorch~\cite{pytorch} on an EC2 instance (with 64 GB NVIDIA T4 Tensor Core GPUs). For feature extraction, we utilize the SAM-Med2D pretrained encoder. Initially, we use annotations for only 10\% of the training dataset, during which all components (visual encoder, prompt encoder, and mask decoder) are fine-tuned. In this step, only the unsupervised prompting strategy is employed. Bounding box and point prompts are used together for all experiments. The remaining portion of the dataset was used in an unannotated form to train the DPO-driven alignment strategy. We optimize the model using the Adam optimizer~\cite{adam}, training for 15 epochs for prompt module fine-tuning and 30 epochs for alignment. In both stages, the initial learning rate is set to 1e-4 and is halved every 10 epochs. All images are resized to a resolution of 256 $\times$ 256 using the same resizing strategy as SAM-Med2D. The loss function used during the initial fine-tuning is a 20:1 weighted combination of focal loss~\cite{focal} and Dice loss~\cite{dice}. While incorporating the preference alignment module, we use the Intersection over Union (IoU) scores between the predicted masks and the ground truth to generate ratings and/or rankings. The IoU scores are binned into the following ranges: \{$<$0.4, 0.4-0.55, 0.55-0.7, and $>$0.7\}. For the generation of multiple segmentation proposals, the model's output probabilities are thresholded at 0.3, 0.4, 0.5, and 0.6. The loss function for training this second stage is listed in Eqn.~\ref{eq:final_model}. The weights $\beta_1=1$ and $\beta_2=0.5$ were experimentally determined to be optimal (see supplementary). Dice Similarity
Coefficient, Intersection over Union (IoU), and Surface Dice Similarity (SDC) scores are used to evaluate segmentation performance.

\begin{figure*}[hbtp]
  \centering
  \includegraphics[width=1.0\linewidth]{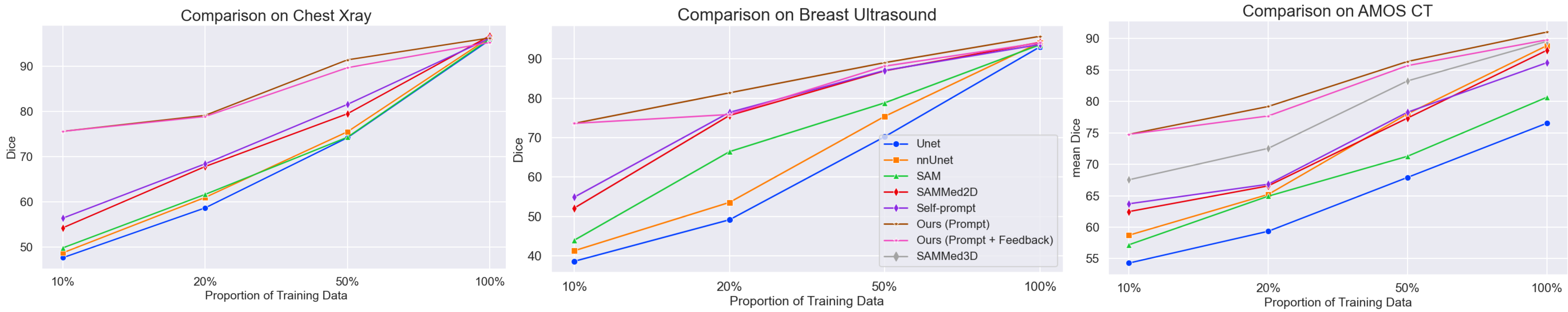}
    \caption{Quantitative comparison with SOTA. Dice score (for Chest Xray, Breast USD) and mean Dice score (for AMOS CT) have been shown to measure the model segmentation performance on different proportions of training data (10\%, 20\%, 50\%, and 100\%).}
    \label{sota}
\end{figure*}


\begin{figure*}[hbtp]
  \centering
\includegraphics[width=0.9\linewidth]{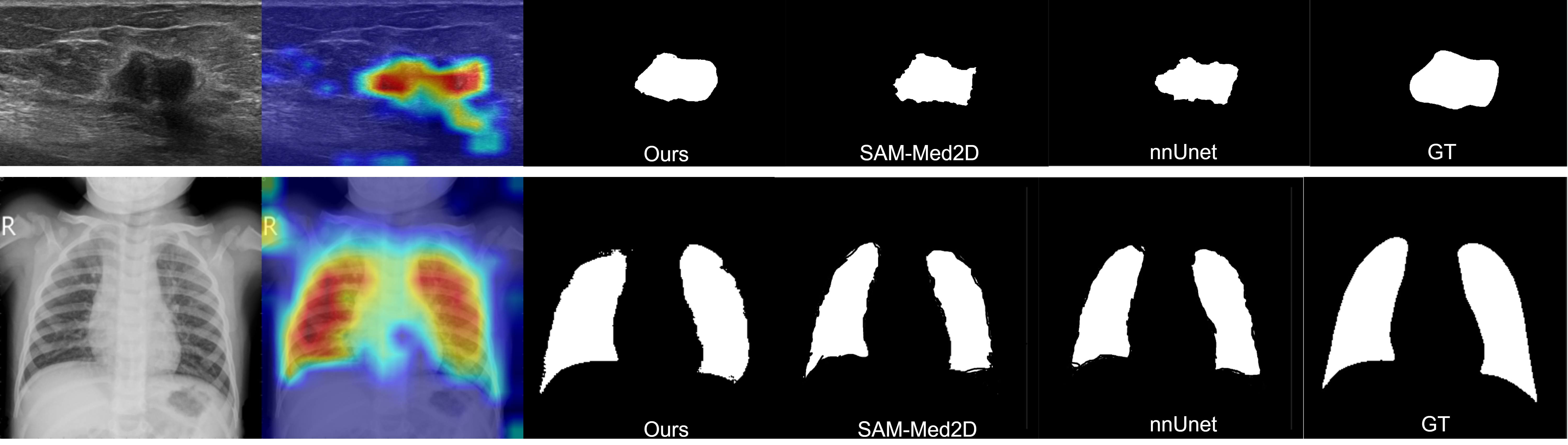}
    \caption{Qualitative comparisons were made between the segmentation results of nnUnet, SAM-Med2D, and our framework on 2D datasets. BiomedCLIP-based saliency maps are also depicted. Experiments were conducted in 50\% data settings.}
    \label{results}
\end{figure*}

\subsection{Comparison with state-of-the-art}
\textbf{Quantitative results.} Fig.~\ref{sota} compares the performance of our framework with relevant methods (U-Net~\cite{unet}, nnU-Net~\cite{nnunet}, SAM~\cite{sam}, SAM-Med2D~\cite{sammed2d}, Self-prompt). The self-prompt method is designed as a variant of ~\cite{samsp}, excluding the knowledge distillation module. We also design a prompt-only baseline of our framework which is trained on different splits of fully annotated data. Our framework is initially trained with only the prompting module using 10\% of data (annotated). As a result, both the prompt-only and final versions exhibit identical performance on this 10\% subset. For the final model, the additional data used for training are considered unannotated, as training the alignment mechanism does not use supervision from ground truth. SOTA methods directly use ground truth for the entire data subset. Nevertheless, our method consistently outperforms them in limited data settings (10-50\%) due to the preference alignment mechanism. Our architecture thus reduces the reliance on large datasets, highlighting its lower annotation requirements, which makes it significantly more effective in low-data regimes. On the Chest-Xray dataset, for instance, with just 20\% of the data, our method achieves a Dice score of 78.87, compared to 58.66 (U-Net), 60.97 (nnU-Net), 61.64 (SAM), 67.81 (SAM-Med2D), and 68.41 (Self-prompt). Both U-Net and nnU-Net are known to be data-hungry models, struggling with smaller data subsets. Our usage of visual and textual prompts, derived through BiomedCLIP and VLM models, offers superior signals to those used in SAM and SAM-Med2D. Unlike these SAM variants, Self-Prompt SAM does not require expert-supplied prompts during inference; it instead generates prompts from the output masks in each iteration. This method offers a slight improvement (+1\%) over SAM-Med2D. Our method shows more stable performance gains across different data subsets, while other SOTA methods exhibit steeper improvements. In the 50\% data setting, our semi-supervised framework achieves an impressive Dice score of 89.68, compared to 91.42 for the supervised prompt-only version.
Similar result trends are observed in the breast ultrasound dataset. Notably, the performance jump of our method from 20\% to 50\% data is much more pronounced than from 10\% to 20\%. This can be due to the nature of the dataset, as the model requires more data to achieve precise and accurate segmentation of breast tumors. We also evaluated our method on a 3D abdominal organ segmentation dataset, including organs such as the liver, kidneys, spleen, pancreas, aorta, bladder, etc (see Fig.~\ref{sota}). Our method outperforms the SOTA across all data proportions except for the full dataset. With 20\% of the data, our method achieves a mean Dice score of 77.69, surpassing U-Net (59.35), nnU-Net (65.21), SAM (64.93), SAM-Med2D (66.57), SAMMed3D (72.54), and Self-prompt (71.83). At the 50\% data setting, it reaches a Dice score of 85.70, comparable to the 86.36 achieved by the prompt-only version.
\begin{table*}[!h]
\centering
\resizebox{0.8\textwidth}{!}{
\begin{tabular}{
>{\columncolor[HTML]{FFFFFF}}c |
>{\columncolor[HTML]{FFFFFF}}c |
>{\columncolor[HTML]{FFFFFF}}c 
>{\columncolor[HTML]{FFFFFF}}c |
>{\columncolor[HTML]{FFFFFF}}c 
>{\columncolor[HTML]{FFFFFF}}c |
>{\columncolor[HTML]{FFFFFF}}c 
>{\columncolor[HTML]{FFFFFF}}c }
\hline
\cellcolor[HTML]{FFFFFF}{\color[HTML]{000000} }                                       & \cellcolor[HTML]{FFFFFF}{\color[HTML]{000000} }                                   & \multicolumn{2}{c|}{\cellcolor[HTML]{FFFFFF}{\color[HTML]{000000} \textbf{Chest-Xray}}} & \multicolumn{2}{c|}{\cellcolor[HTML]{FFFFFF}{\color[HTML]{000000} \textbf{Breast-USD}}} & \multicolumn{2}{c}{\cellcolor[HTML]{FFFFFF}{\color[HTML]{000000} \textbf{AMOS-CT}}} \\ \cline{3-8} 
\multirow{-2}{*}{\cellcolor[HTML]{FFFFFF}{\color[HTML]{000000} \textbf{Supervision}}} & \multirow{-2}{*}{\cellcolor[HTML]{FFFFFF}{\color[HTML]{000000} \textbf{Methods}}} & {\color[HTML]{000000} \textbf{IoU}}        & {\color[HTML]{000000} \textbf{Dice}}       & {\color[HTML]{000000} \textbf{IoU}}        & {\color[HTML]{000000} \textbf{Dice}}       & {\color[HTML]{000000} \textbf{mDice}}     & {\color[HTML]{000000} \textbf{mSDC}}      \\ \hline
10\% + 10\% unannotated                                                                 & Ours                                                                              & {\color[HTML]{FE0000} \textbf{74.30}}      & {\color[HTML]{FE0000} \textbf{78.87}}      & {\color[HTML]{FE0000} \textbf{64.35}}      & {\color[HTML]{FE0000} \textbf{75.88}}      & {\color[HTML]{FE0000} \textbf{77.69}}    & {\color[HTML]{FE0000} \textbf{78.34}}    \\ \hline
20\%                                                                                  & - Alignment                                                                        & \textbf{75.02}                             & \textbf{79.13}                             & \textbf{67.51}                             & \textbf{81.38}                             & \textbf{79.20}                           & \textbf{80.66}                           \\ \hline
\cellcolor[HTML]{FFFFFF}{\color[HTML]{000000} }                                       & {\color[HTML]{000000} - Alignment}                                                 & {\color[HTML]{000000} 68.43}               & {\color[HTML]{000000} 75.60}               & {\color[HTML]{000000} 61.44}               & {\color[HTML]{000000} 73.62}               & {\color[HTML]{000000} 74.77}             & {\color[HTML]{000000} 76.06}             \\
\cellcolor[HTML]{FFFFFF}{\color[HTML]{000000} }                                       & {\color[HTML]{000000} - Alignment - VQA}                                           & {\color[HTML]{000000} 66.90}               & {\color[HTML]{000000} 73.35}               & {\color[HTML]{000000} 57.60}               & {\color[HTML]{000000} 70.53}               & {\color[HTML]{000000} 74.08}             & {\color[HTML]{000000} 75.41}             \\
\cellcolor[HTML]{FFFFFF}{\color[HTML]{000000} }                                       & {\color[HTML]{000000} - Alignment - VQA - GPT4}                                    & {\color[HTML]{000000} 63.16}               & {\color[HTML]{000000} 72.76}               & {\color[HTML]{000000} 54.26}               & {\color[HTML]{000000} 68.89}               & {\color[HTML]{000000} 73.16}             & {\color[HTML]{000000} 74.89}             \\
\multirow{-4}{*}{\cellcolor[HTML]{FFFFFF}{\color[HTML]{000000} 10\%}}                 & {\color[HTML]{000000} - Alignment - VQA - CAM}                                     & {\color[HTML]{000000} 50.38}               & {\color[HTML]{000000} 57.02}               & {\color[HTML]{000000} 45.16}               & {\color[HTML]{000000} 59.05}               & {\color[HTML]{000000} 69.97}             & {\color[HTML]{000000} 71.45}             \\ \hline
\end{tabular}}
\caption{Ablation results demonstrating
effectiveness of major components.}
    \label{ablation1}
\end{table*}

\begin{table*}[!h]
\centering
\resizebox{0.8\textwidth}{!}{
\begin{tabular}{
>{\columncolor[HTML]{FFFFFF}}c |
>{\columncolor[HTML]{FFFFFF}}c |
>{\columncolor[HTML]{FFFFFF}}c 
>{\columncolor[HTML]{FFFFFF}}c |
>{\columncolor[HTML]{FFFFFF}}c 
>{\columncolor[HTML]{FFFFFF}}c |
>{\columncolor[HTML]{FFFFFF}}c 
>{\columncolor[HTML]{FFFFFF}}c }
\hline
\cellcolor[HTML]{FFFFFF}{\color[HTML]{000000} }                                       & \cellcolor[HTML]{FFFFFF}{\color[HTML]{000000} }                                       & \multicolumn{2}{c|}{\cellcolor[HTML]{FFFFFF}{\color[HTML]{000000} \textbf{Chest-Xray}}} & \multicolumn{2}{c|}{\cellcolor[HTML]{FFFFFF}{\color[HTML]{000000} \textbf{Breast-USD}}} & \multicolumn{2}{c}{\cellcolor[HTML]{FFFFFF}{\color[HTML]{000000} \textbf{AMOS-CT}}} \\ \cline{3-8} 
\multirow{-2}{*}{\cellcolor[HTML]{FFFFFF}{\color[HTML]{000000} \textbf{Supervision}}} & \multirow{-2}{*}{\cellcolor[HTML]{FFFFFF}{\color[HTML]{000000} \textbf{Preference Alignment}}} & {\color[HTML]{000000} \textbf{IoU}}        & {\color[HTML]{000000} \textbf{Dice}}       & {\color[HTML]{000000} \textbf{IoU}}        & {\color[HTML]{000000} \textbf{Dice}}       & {\color[HTML]{000000} \textbf{mDice}}      & {\color[HTML]{000000} \textbf{mSDC}}     \\ \hline
\cellcolor[HTML]{FFFFFF}                                                              & Loss for best candidate                                                               & {\color[HTML]{000000} 70.37}               & {\color[HTML]{000000} 77.09}               & {\color[HTML]{000000} 61.76}               & {\color[HTML]{000000} 73.81}               & {\color[HTML]{000000} 75.01}              & {\color[HTML]{000000} 76.36}            \\
\cellcolor[HTML]{FFFFFF}                                                              & Rating                                                                                & 73.99                                      & 78.41                                      & 63.97                                      & 75.52                                      & 77.23                                     & 77.98                                   \\
\multirow{-3}{*}{\cellcolor[HTML]{FFFFFF}10\% + 10\% unannotated}                       & {\color[HTML]{000000} Ranking}                                                        & {\color[HTML]{000000} 74.30}               & {\color[HTML]{000000} 78.87}               & {\color[HTML]{000000} 64.35}               & {\color[HTML]{000000} 75.88}               & {\color[HTML]{000000} 77.69}              & {\color[HTML]{000000} 78.34}            \\ \hline
\cellcolor[HTML]{FFFFFF}{\color[HTML]{000000} 10\% + 20\% unannotated}                  & {\color[HTML]{000000} Ranking}                                                        & {\color[HTML]{000000} 79.74}               & {\color[HTML]{000000} 85.15}               & {\color[HTML]{000000} 73.56}               & {\color[HTML]{000000} 84.23}               & {\color[HTML]{000000} 80.54}              & {\color[HTML]{000000} 81.95}            \\ \cline{1-1}
\cellcolor[HTML]{FFFFFF}{\color[HTML]{000000} 10\% + 40\% unannotated}                  & {\color[HTML]{000000} Ranking}                                                        & {\color[HTML]{000000} 88.96}               & {\color[HTML]{000000} 89.68}               & {\color[HTML]{000000} 85.92}               & {\color[HTML]{000000} 88.15}               & {\color[HTML]{000000} 84.30}              & {\color[HTML]{000000} 85.70}            \\ \hline
\end{tabular}}
\caption{Ablation results on different proportions of training data for 3 types of preference scoring strategies.}
    \label{ablation2}
\end{table*}

\begin{figure}[hbtp]
  \centering
\includegraphics[width=0.9\linewidth]{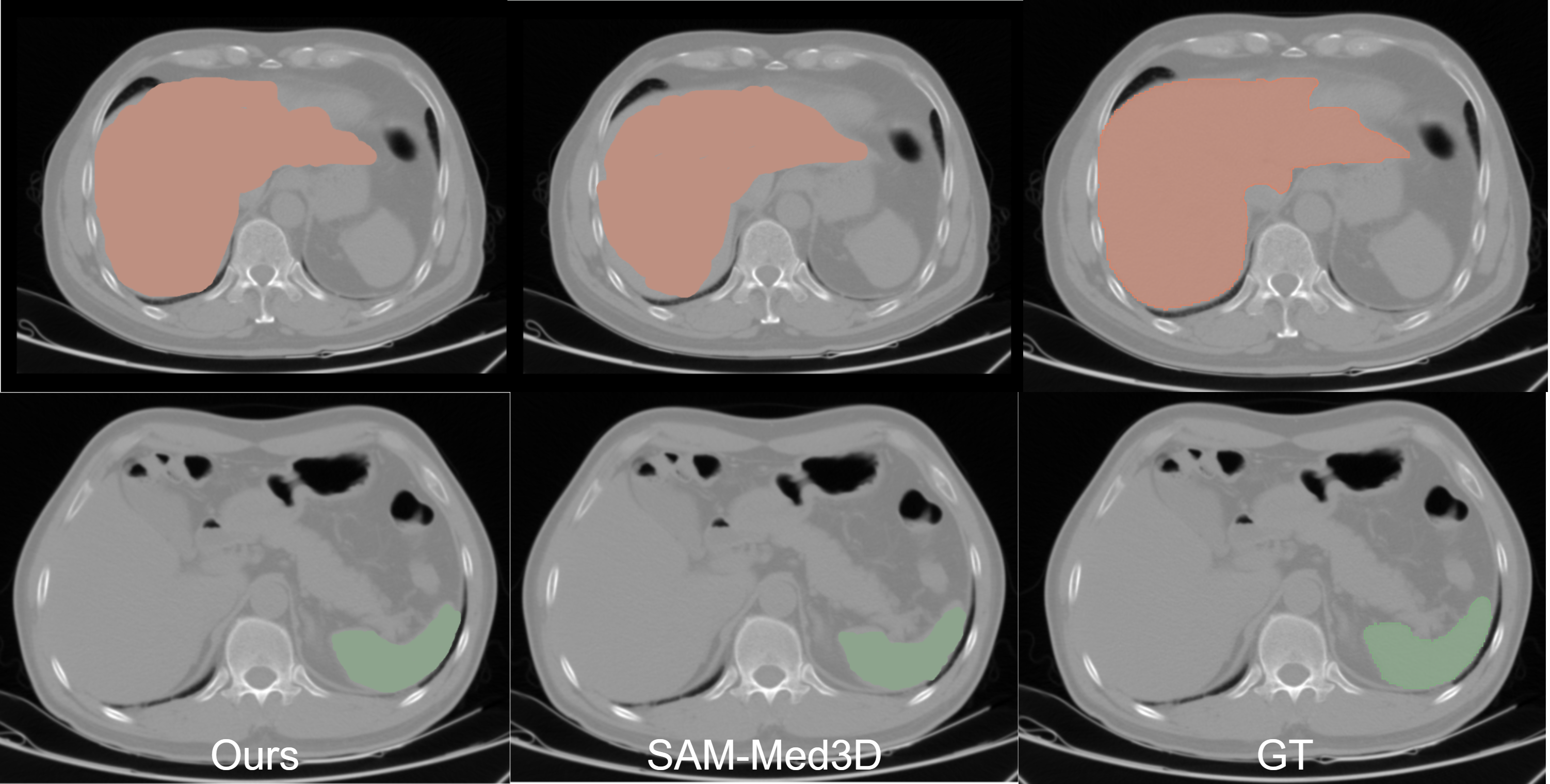}
    \caption{Segmentation maps of different anatomical structures (liver and spleen) for SAM-Med3D and our method}
    \label{results2}
\end{figure}

\textbf{Qualitative results.}
We present segmentation maps for 3 datasets in Fig.~\ref{results} and Fig.~\ref{results2}, generated by our model trained at 50\% data settings. The saliency maps could correctly highlight the target regions and proved to be an excellent source of supervision. It can be noted from Fig.~\ref{results} that our segmentation quality around the boundaries of tumor or lung is much superior compared to both nnUnet and SAM-Med2D. In Fig.~\ref{results2}, SAM-Med3D tended to under-segment the edges of both the abdominal organs, the liver, and spleen. More results provided in the supplementary.
 
\subsection{Ablation studies}
\textbf{Effectiveness of major components.} We conduct several ablation experiments (see Tab.~\ref{ablation1}) to evaluate the contribution of each module in our architecture. For simplicity, we focus on the Chest-Xray dataset for analysis. Before incorporating the DPO-driven alignment strategy, all baselines were trained with 10\% labeled data. First, we developed a naive baseline (last row), which relies solely on textual answers generated from GPT-4. As expected, it performed poorly, achieving a 57.02 Dice score. A second baseline (second-last row) was designed to use only visual prompts from BiomedCLIP, which provides semantic information. This method significantly benefited from the coarse segmentation masks derived from CLIP saliency maps, improving the Dice score to 72.76. Next, we combined both textual and visual prompts to form a third baseline (-Alignment-VQA), which resulted in a slight improvement (+0.59\%) over the CLIP-only baseline. Finally, we integrated VQA into the prompting strategy to obtain answers related to the shape and location of the target regions, completing our fully empowered unsupervised prompting strategy (-Alignment). This model achieved a Dice score of 75.60 with 10\% data and improved to 79.13 with 20\% data. In comparison, with the preference alignment mechanism, fine-tuning the prompting module with 10\% annotated data and training the alignment module on an additional 10\% unannotated data achieved a Dice score of 78.87. This is on par with the fully supervised prompt-only model, underscoring the effectiveness of our alignment module.

\textbf{Preference Scoring strategies.}
We conducted several ablation studies (detailed in Tab.~\ref{ablation2}) to evaluate the effectiveness of the preference-scoring strategy. One baseline approach involved backpropagating the loss based only on the best candidate, while another compared ranking the candidates rather than simply rating them. The results are presented for different proportions of unannotated data, on top of the foundational fine-tuning of both the prompt encoder and the decoder using 10\% annotated data. At 10\% unannotated data, both the ranking and rating approaches performed similarly (mean Dice scores of 77.69 and 77.23 for AMOS CT, respectively), significantly outperforming the best-candidate-only method (75.01). Additionally, we observed that as the proportion of unannotated data increased, the model's performance improved substantially. 

\textbf{Robustness to noisy rating.} We experimented with introducing noise into the rating mechanism by flipping the ratings of closely ranked candidates to enhance the framework's robustness. Results are in the supplementary.

\section{Conclusion}
We introduce a novel training strategy to enhance SAM representations for semi-supervised medical image segmentation. We extract integrated semantic, location, and shape information from pretrained vision-language models in an unsupervised manner. This information is used as refined prompts for our model. We also implement an optimal policy, inspired by direct preference optimization in language models. This enables human-in-the-loop feedback simulation within a streamlined training framework, without the need for a separate reward function or extensive knowledge from annotators. These modules ensure that our framework outperforms state-of-the-art methods across datasets spanning multiple modalities in low-annotation data scenarios.




{
    \small
    \bibliographystyle{ieeenat_fullname}
    \bibliography{main}
}

\clearpage
\setcounter{page}{1}
\setcounter{section}{0}
\maketitlesupplementary

\setlength{\tabcolsep}{3pt}
\begin{table*}[b!]
\fontsize{7}{9}\selectfont
\centering
\begin{tabular}{c|ccc}
\hline
\multirow{2}{*}{\textbf{VLM}} & \multicolumn{3}{c}{\textbf{Prompts}}                                                                                                                                                                                                                                                                                                                                                                                 \\ \cline{2-4} 
                     & \multicolumn{1}{c|}{\textbf{Chest X-ray}}                                                                                                              & \multicolumn{1}{c|}{\textbf{Breast USD}}                                                                                                                    & \textbf{AMOS-CT}                                                                                                 \\ \hline
BiomedCLIP           & \multicolumn{1}{c|}{chest x-ray}                                                                                                              & \multicolumn{1}{c|}{{[}class{]} breast tumor}                                                                                                      & {[}organ{]}                                                                                             \\ \hline
MedVInT              & \multicolumn{1}{c|}{\begin{tabular}[c]{@{}c@{}}Briefly describe the condition of lungs and \\ location of pathologies\end{tabular}}           & \multicolumn{1}{c|}{\begin{tabular}[c]{@{}c@{}}What is the shape of breast tumor and \\ where is it located?\end{tabular}}                         & \begin{tabular}[c]{@{}c@{}}What is the shape of the {[}organ{]}\\ and where is it located?\end{tabular} \\ \hline
GPT-4                & \multicolumn{1}{c|}{\begin{tabular}[c]{@{}c@{}}Briefly describe, in one line, the lungs\\ of a patient suffering from [disease]\end{tabular}} & \multicolumn{1}{c|}{\begin{tabular}[c]{@{}c@{}}Briefly describe, in one line, {[}class{]} breast \\ tumor of a patient in Ultrasound\end{tabular}} & \begin{tabular}[c]{@{}c@{}}Briefly describe, in one line, {[}organ{]} of \\ a human in CT\end{tabular}  \\ \hline
\end{tabular}
\caption{Different prompts designed for the BiomedCLIP, MedVInT, and GPT-4 models. The placeholder [class] refers to the tumor type, either malignant or benign, while [organ] refers to one of the 15 organs available in the AMOS-CT dataset for segmentation.}\label{tab:design}
\end{table*}

\section{More Qualitative Results}
\begin{figure}[!hbtp]
  \centering
\includegraphics[width=0.9\linewidth]{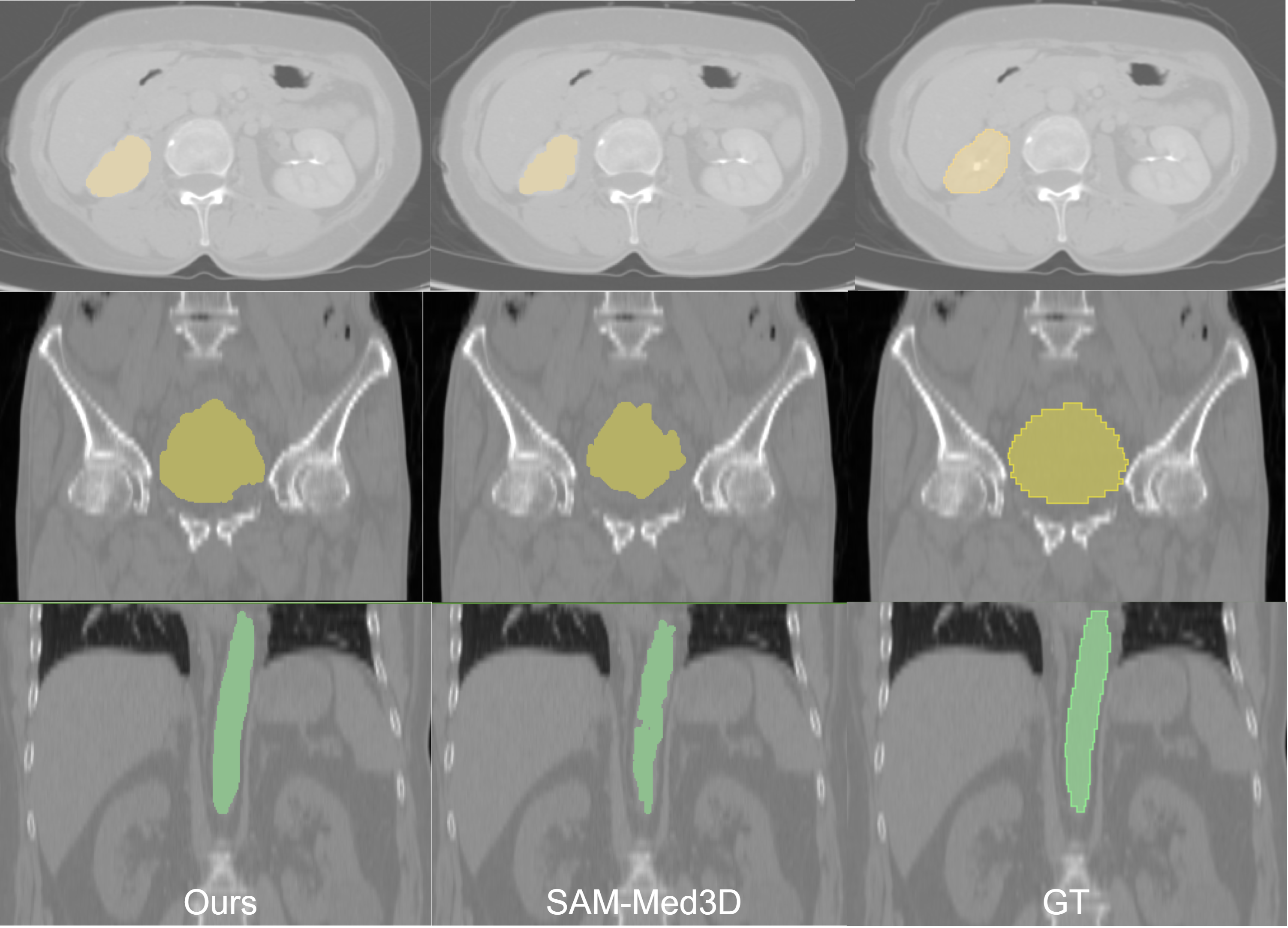}
    \caption{Segmentation maps of three anatomical structures (right kidney, bladder, and aorta)$\downarrow$ for SAM-Med3D and our method}
    \label{sota2_supp}
\end{figure}

\begin{figure*}[!h]
  \centering
\includegraphics[width=0.9\linewidth]{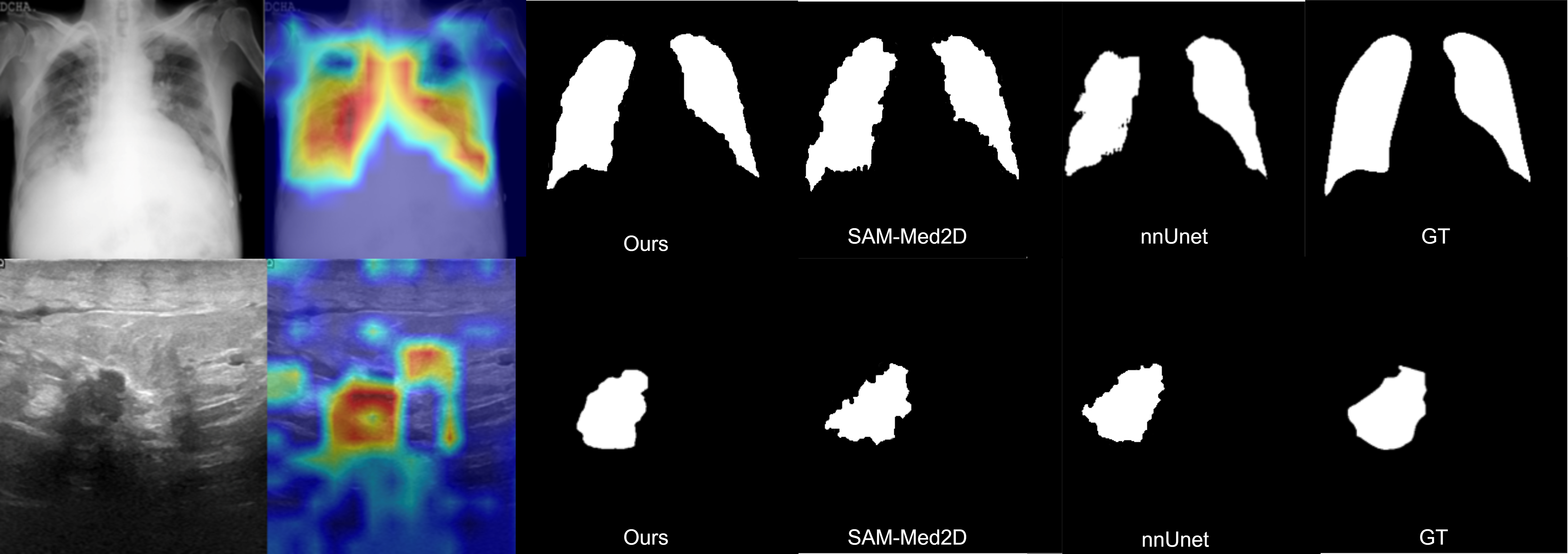}
    \caption{Qualitative comparisons were made between the segmentation results of nnUnet, SAM-Med2D, and our framework on 2D datasets. BiomedCLIP-based saliency maps are also depicted. Experiments were conducted in 50\% data settings.}
    \label{sota_supp}
\end{figure*}

We present segmentation maps for 3 datasets in Fig.~\ref{sota_supp} and Fig.~\ref{sota2_supp}, generated by our model trained at 50\% data settings. The saliency maps could correctly highlight the target regions and proved to be an excellent source of supervision. It can be noted from Fig.~\ref{sota_supp} that our segmentation quality around the boundaries of tumor or lung is much superior compared to both nnUnet and SAM-Med2D. In Fig.~\ref{sota2_supp}, similar trends can be seen while segmenting the abdominal organs -- right kidney, bladder, and aorta (top to bottom).

\section{Robustness to noise in rating}
\label{sec:noise}
We randomly flipped one of three rating combinations (1$\leftrightarrow$2, 2$\leftrightarrow$3,3$\leftrightarrow$4) for 5-30\% of the training image samples. This was done to evaluate the robustness of our framework to noise in the rating process. Despite the introduction of noise through the virtual annotator, the Dice scores showed minimal decline. The results have been shown in Tab.~\ref{tab:noise}. With 30\% of the image samples affected by noisy ratings, the dice score performance decreased by only 0.24, 0.20, and 0.24 for the X-ray, USD, and CT datasets, respectively.

\setlength{\tabcolsep}{3pt}
\begin{table}[!h]
\fontsize{10}{12}\selectfont
\centering
\begin{tabular}{cl|ccc}
\hline
\multicolumn{2}{c|}{\multirow{2}{*}{Flip (\%)}} & \multicolumn{3}{c}{Dice score (20\% data)}                                                 \\ \cline{3-5} 
\multicolumn{2}{c|}{}                           & \multicolumn{1}{c|}{Chest-Xray}     & \multicolumn{1}{c|}{Breast-USD}     & AMOS-CT (mean) \\ \hline
\multicolumn{2}{c|}{\textbf{0}}                 & \multicolumn{1}{c|}{\textbf{78.87}} & \multicolumn{1}{c|}{\textbf{75.88}} & \textbf{77.69} \\
\multicolumn{2}{c|}{5}                          & \multicolumn{1}{c|}{78.82}          & \multicolumn{1}{c|}{75.83}          & 77.62          \\
\multicolumn{2}{c|}{10}                         & \multicolumn{1}{c|}{78.79}          & \multicolumn{1}{c|}{75.81}          & 77.58          \\
\multicolumn{2}{c|}{20}                         & \multicolumn{1}{c|}{78.71}          & \multicolumn{1}{c|}{75.74}          & 77.51          \\
\multicolumn{2}{c|}{30}                         & \multicolumn{1}{c|}{78.63}          & \multicolumn{1}{c|}{75.68}          & 77.45          \\ \hline
\end{tabular}
\caption{Ablation results for varying proportions (5\%-30\%) of images with flipped ratings.}\label{tab:noise}
\end{table}

\section{Selection of experimental parameters $\beta_1$, $\beta_2$}
\label{sec:beta}
We tested different pairs of $\beta_1$, and $\beta_2$ values to identify the optimal combination in Eqn. 5
.
As shown in Tab.~\ref{parameter}, the best performance was achieved with  $\beta_1=1$, and $\beta_2=0.5$.

\setlength{\tabcolsep}{2pt}
\begin{table}[!h]
\fontsize{11}{14}\selectfont
\centering
\begin{tabular}{c|c|ccc}
\hline
\multirow{2}{*}{$\beta_1$} & \multirow{2}{*}{$\beta_2$} & \multicolumn{3}{c}{Dice score (20\% data)}                                                 \\ \cline{3-5} 
                           &                            & \multicolumn{1}{c|}{Chest-Xray}     & \multicolumn{1}{c|}{Breast-USD}     & AMOS-CT (mean) \\ \hline
2                          & 1                          & \multicolumn{1}{c|}{78.12}          & \multicolumn{1}{c|}{75.43}          & 77.47          \\
1.5                        & 0.75                       & \multicolumn{1}{c|}{78.64}          & \multicolumn{1}{c|}{75.70}          & 77.53          \\
\textbf{1}                 & \textbf{0.5}               & \multicolumn{1}{c|}{\textbf{78.87}} & \multicolumn{1}{c|}{\textbf{75.88}} & \textbf{77.69} \\ \hline
\end{tabular}
\caption{Selection of experimental parameters $\beta_1$, $\beta_2$}\label{parameter}
\end{table}

\section{Prompt design}\label{sec:prompt}
Text-based prompts were designed to provide inputs for the BiomedCLIP, MedVInT, and GPT-4 models, enabling both direct and indirect supervision from them. This supervision can take the form of responses or guidance for generating saliency maps. A summary of the design for each of the three datasets is provided in the Tab.~\ref{tab:design}.

\end{document}